\documentclass[letterpaper]{article}

\usepackage{natbib,alifeconf}  
\usepackage{amsmath,amsfonts,amssymb}
\usepackage{url,hyperref,cleveref}
\usepackage{booktabs}
\usepackage{algorithm}
\usepackage{algorithmic}
\usepackage{multirow}
\usepackage{enumitem}

\hypersetup{
    colorlinks=true,
    linkcolor=blue,
    urlcolor=blue,
    citecolor=blue
}

\usepackage{xcolor}
\usepackage[normalem]{ulem}
\usepackage{soul}
\definecolor{newhl}{rgb}{1,1,0.45}
\sethlcolor{newhl}
\newif\iftrackchanges
\trackchangesfalse
\iftrackchanges
  \newcommand{\old}[1]{{\color{red}\sout{#1}}}
  \newcommand{\new}[1]{\hl{#1}}
\else
  \newcommand{\old}[1]{}
  \newcommand{\new}[1]{#1}
\fi

\newcommand\blfootnote[1]{%
  \begingroup
  \renewcommand\thefootnote{}\footnote{#1}%
  \addtocounter{footnote}{-1}%
  \endgroup
}

\title{Architecture Generalization with MetaNCA}

\author{Meet Barot*$^{1}$, Daniel Berenberg$^{1}$ \and Sina Khajehabdollahi$^2$ \\
\mbox{}\\
$^1$Mythos Scientific, New York, USA \\
$^2$Independent Scholar\\
\texttt{*meet@mythos.science}}
\begin{document}

\maketitle

\begin{abstract}
	Self-organization is an emergent property of life, driven by the collective behavior of individual components acting on local information.
	Biological neurons, through local interactions transmitted through synapses, are able to learn efficiently and can adapt their connections over an organism's lifespan.
	Motivated by these desirable properties of adaptability and local interaction, neural cellular automata (NCA) models have been successful at learning morphogenesis solely through local update rules, demonstrating stability over many updates and robustness to perturbations\old{\mbox{ \citep{mordvintsev2020growing}}}.
	In this work, we introduce Meta Neural Cellular Automata (MetaNCA), a framework that learns local rules which self-organize the weights of artificial neural networks.
    A learned rule network iteratively updates the weights of a task network using only local interactions on the computation graph.
	We propose a novel Weight Transformer architecture for the local rule network, which uses linear attention to aggregate signals from neighboring weights and hidden states.
        Once trained, the rule network generates task networks of \old{arbitrary}\new{diverse} architectures without backpropagation.
    We show that MetaNCA generates weights for feedforward MLPs, CNNs, and ResNets on MNIST and CIFAR-100, scaling to networks of 2 million parameters.
	We further show that MetaNCA generalizes to architectures not seen during meta-training, and that architectural diversity in the training phase strengthens this generalization.
\end{abstract}

Data/Code available at: \url{https://github.com/Mythos-Scientific/meta-nca}
\blfootnote{\textcopyright  2026 Barot, Berenberg \& Khajehabdollahi. Published under a Creative Commons Attribution 4.0 International (CC BY 4.0) license.}

\section{Introduction}
Backpropagation and gradient-based optimization methods for training neural networks are responsible for much of the impressive capabilities of deep learning models when applied to large datasets. However, many drawbacks to these methods exist: the memory consumption of gradient calculations, the requirement of an unchanging architecture in the training process, and the necessity of a large number of examples to learn new concepts. There exists a stark contrast between the memory and power costs for training large models compared to their analogs in nature (e.g. the mammalian brain), which consume orders of magnitude less power and are subject to a genomic bottleneck that strongly limits the level of description required to develop them. Notably, the information capacity of the genome is orders of magnitude smaller than the complexity of neural circuitry \citep{zador2019critique, shuvaev2024encoding}. This implies that what is actually encoded are more generalized rules for developing and growing the circuitry rather than the specific details of the circuitry itself. The biological analogy points toward a different path forward, where the goal shifts from scaling the specifics of large models to learning the rules that generate them.

As the demand to scale models continues to grow, this gap between current machine learning methods and biological systems becomes increasingly difficult to ignore. These resource constraints affect not just the training or inference of these models, but also their flexibility and adaptability when it comes to fine-tuning and specialization, where the need to scale costs and dimensionality down have inspired methods such as LoRA \citep{hu2022lora} that operate in lower rank spaces. There is therefore an increasing demand for more universal and highly compressible models that challenge the existing paradigms of training large foundation models.

Existing approaches each address different parts of this challenge, approaching the challenge from different traditions. Traditional hypernetworks and weight-space methods can generate neural network parameters from compact representations, but typically operate globally when generating weight matrices for a fixed target architecture \citep{ha2016hypernetworks, chauhan2024brief}. Learned plasticity rules offer local, biologically inspired updates, though cross-architecture generalization has been limited in scope \citep{miconi2018differentiable, miconi2020backpropamine}. HyperNCA \citep{najarro2022hypernca} and neural developmental programs \citep{najarro2023towards, plantec2024evolving} have shown that NCAs can generate network parameters through local rules, but operate on fixed grid structures or grow topology from scratch rather than generating weights for \old{arbitrary}\new{a range of} pre-specified architectures. To our knowledge, no existing method combines purely local weight updates on the computational graph with training across a distribution of architectures via end-to-end backpropagation.

To this end, we introduce MetaNCA, a neural cellular automaton that generates ensembles of large neural networks of various architectures through local self-organization. MetaNCAs are graph neural cellular automata that iteratively generate the parameters of a model using only the local information of an edge in the model's computational graph. \new{Because this local rule is defined purely from a weight's forward and backward neighbors on the computation graph (the edges sharing its output and input nodes), it is by construction not tied to any particular architecture.} During training, MetaNCA must generate parameters for a variety of architectures concurrently to ensure that parameter-generating rules do not overfit to a particular configuration. To our knowledge, MetaNCA is the first method to learn a single local rule that can self-organize across a distribution of unseen architectures with millions of parameters. We demonstrate these capabilities on MLPs, CNNs and ResNets.

\begin{figure}
  \centering	\includegraphics[width=0.5\textwidth]{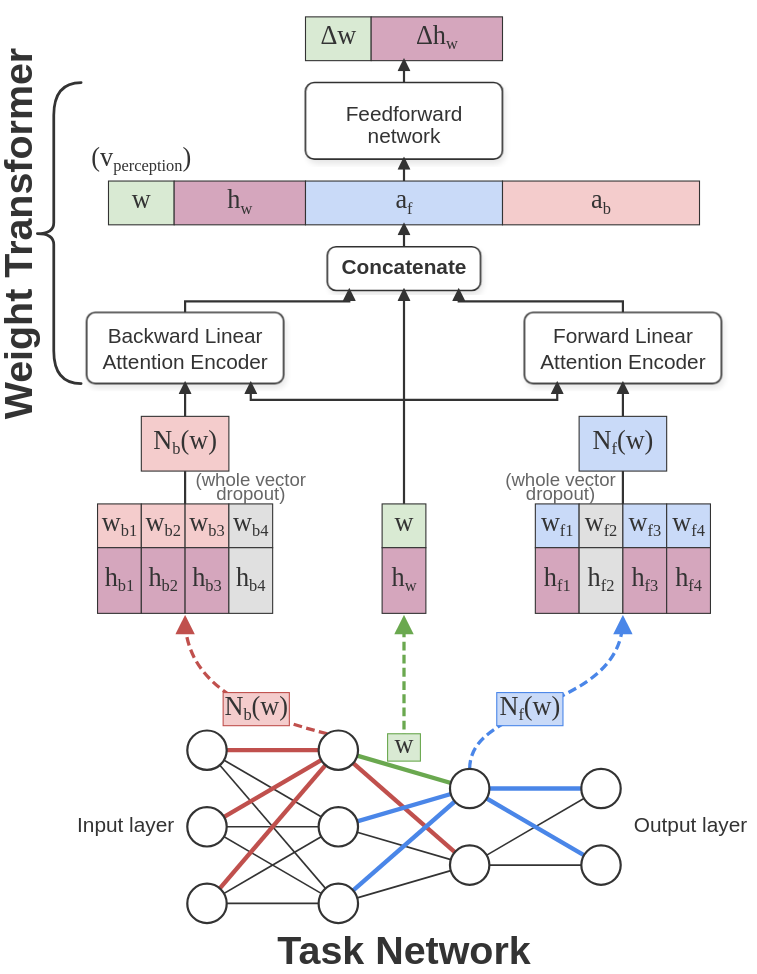}
	\caption{Diagram of the Weight Transformer local rule network producing an update to an example weight $w$ in an MLP task network. The input to the local rule includes $w$ and its forward and backward neighborhoods $\mathcal{N}_{f}$ and $\mathcal{N}_{b}$, and their corresponding hidden states. Cross attention outputs are given by $a_{f}$ and $a_{b}$.}
	\label{fig:pipeline}
\end{figure}

\subsection{Related Work}
Our methods are inspired by a growing body of research that has worked on bridging the gap between complex dynamical systems, self-organization, and deep learning.

\subsubsection{Plasticity}
Plasticity-based learning algorithms are biologically motivated methods to self-organize neural networks using local learning rules. They operate by modifying the synaptic weights of a network as a function of its local activity $\Delta w_{ij} = f(a_{\{i, j\}}, w_{ij})$. While plasticity-based methods are more biologically grounded, they often fail to compete with modern artificial neural networks that do not bother with biological realism (e.g. the backpropagation algorithm). More recently, work to bridge these limitations and apply more modern optimization techniques to these local learning rules have been made \citep{miconi2018differentiable, miconi2020backpropamine, najarro2020meta}, pushing the capabilities of these local learning models. \new{Earlier work evolved a neural network to serve as the learning rule itself }\citep{runarsson2000evolution}\new{, albeit demonstrated only on single-layer networks.} However, to date, these methods show limited generalization to unseen architectures.

\subsubsection{Neural Cellular Automata}
The conceptual roots of this work trace back to cellular automata (CA): simple rule-based systems in which complex global behavior emerges purely from local interactions. From elementary 1D CAs\new{\mbox{ \citep{wolfram1983statistical}}} to Conway's Game of Life\new{\mbox{ \citep{gardner1970mathematical}}}, these systems demonstrated that remarkably rich structure, including self-replication, pattern formation, and universal computation, can arise from a handful of local rules applied iteratively on a grid.

Growing Neural Cellular Automata (NCA) \citep{mordvintsev2020growing} formalized this intuition for the deep learning era, replacing hand-crafted rules with learned local update functions operating on grids or graphs, producing systems reminiscent of reaction-diffusion models. These systems proved surprisingly capable: NCAs have since been trained to grow and robustly regenerate target patterns from a single seed cell \citep{mordvintsev2020growing}, synthesize textures \citep{niklasson2021self-organising}, classify MNIST digits \citep{randazzo2020self-classifying} all without any global coordinator, the competence emerging entirely from the local rule. These methods have also been extended to generalize to broader distributions of tasks \citep{guichard2025engramnca, guichard2025arc}.

\subsubsection{Hypernetworks}
While NCAs have demonstrated remarkable success at tasks like pattern recognition and classification through purely local computation, we ask whether this same paradigm can be turned toward a more meta objective: generating the parameters of a neural network itself. This connects to the growing literature on weight-space learning \citep{han2026survey}: methods that operate on or generate neural network parameters directly. Hypernetworks \citep{stanley2009hypercube, ha2016hypernetworks, chauhan2024brief} learn to produce the weights of a larger network from a compact model and have been extended for continual learning \citep{von2019continual} and architecture search \citep{brock2017smash}. More recently, diffusion-based approaches generate network parameters by learning distributions over weight spaces \citep{wang2024neural}, and hyper-representations learn reusable embeddings of trained network checkpoints \citep{schurholt2022hyper}. \old{However, none of these methods operate at the level of individual weights. Instead, they generate parameters at the granularity of layers, chunks, or parameter subsets, coupling the generator to a specific target architecture. MetaNCA instead applies its local rule independently to each weight in the computational graph, which decouples it from any particular architecture and enables cross-architecture generalization by construction.}
\new{Many of these methods generate parameters at the granularity of layers, chunks, or parameter subsets, coupling the generator to a specific target architecture. Indirect encodings such as HyperNEAT }\citep{stanley2009hypercube}\new{ are a notable exception, querying a CPPN with the geometric coordinates of each connection to produce individual weights; however, this is a single static query of a coordinate function rather than an iterative update rule, and it does not aggregate information from a weight's neighborhood in the computation graph. MetaNCA instead applies an iterative local update rule to each weight using its forward and backward neighborhoods on the computation graph, and meta-trains this rule across a distribution of architectures, which decouples it from any particular architecture and enables cross-architecture generalization.}

The most direct precursor to our work is HyperNCA \citep{najarro2022hypernca}, which first combined Hypernetworks and NCAs by using an NCA as the weight-generating mechanism itself: a compact local rule iteratively develops a seed state whose grown pattern is read out as the weights of a policy network. This developmental paradigm has since been extended through Neural Developmental Programs \citep{najarro2023towards}, which grow network structure and weights via local inter-node communication alone, and further toward self-assembling networks with structural and synaptic plasticity \citep{plantec2024evolving}. \old{MetaNCA builds on this line of work by explicitly training the local rule to generate weight updates for multiple fixed architectures, resulting in generalization to unseen architectures.}\new{MetaNCA builds on this line of work but differs from these developmental approaches in a key respect: rather than growing structure or reading out weights from a separate substrate or grid, its local rule operates directly on the edges of the target computation graph, updating each weight from its forward and backward neighborhoods. By explicitly training this rule across multiple architectures, MetaNCA generalizes to architectures unseen during meta-training.}

\section{Method Description}

In MetaNCA, a local rule is applied to each individual weight and uses information in its local neighborhood to produce updates. A linear attention block generates local perception vectors which feeds into an MLP that updates the local weight and hidden state. This process iterates for $T=10$ timesteps. As an analogy to the Growing NCA \citep{mordvintsev2020growing} work, in which a local rule neural network applies updates to pixels on a grid, MetaNCA operates on the weight values as \old{"pixels"}\new{\mbox{``pixels''}}, with the \old{"grid"}\new{\mbox{``grid''}} being the neural network graph.

\paragraph{Problem setup.}
We have a dataset $D$ of samples $(x, y)$ representing the inputs and labels respectively for a classification task, and a corresponding task neural network $\mathcal{T}$ that makes predictions for this task $\hat{y} = \mathcal{T}(\theta; x)$.
The task network $\mathcal{T}$ has weights $w^{L}_{ij} \in \theta$, where $L$ is a layer of the task network and $i$ and $j$ are indices of the weight matrix for layer $L$.
In addition, for each weight $w_{ij}$, the task network also has a hidden state vector $\vec{h}_{ij}$\new{: a per-weight latent vector, carried alongside the weight across update steps, that lets the local rule store and communicate information beyond the scalar weight value}. In practice we use many task networks in order to train the local rule network to generalize across architectures, but for simplicity we will describe MetaNCA assuming a single task network.

A separate local rule neural network $R$, parametrized by $\phi$, updates $\theta^{(t)}$ and $H^{(t)}$\new{, where $\theta^{(t)}$ denotes the full set of task-network weights and $H^{(t)} = \{\vec{h}_{ij}\}$ the corresponding set of all hidden states, both at update step $t$}.
For a given weight $w_{ij}$, we take the neighboring weight and hidden state vectors of $w_{ij}$ as the input to the local rule network R.

\subsection{Local rule network inputs.}

\begin{figure*}[ht!]
    \centering
    \includegraphics[width=\textwidth]{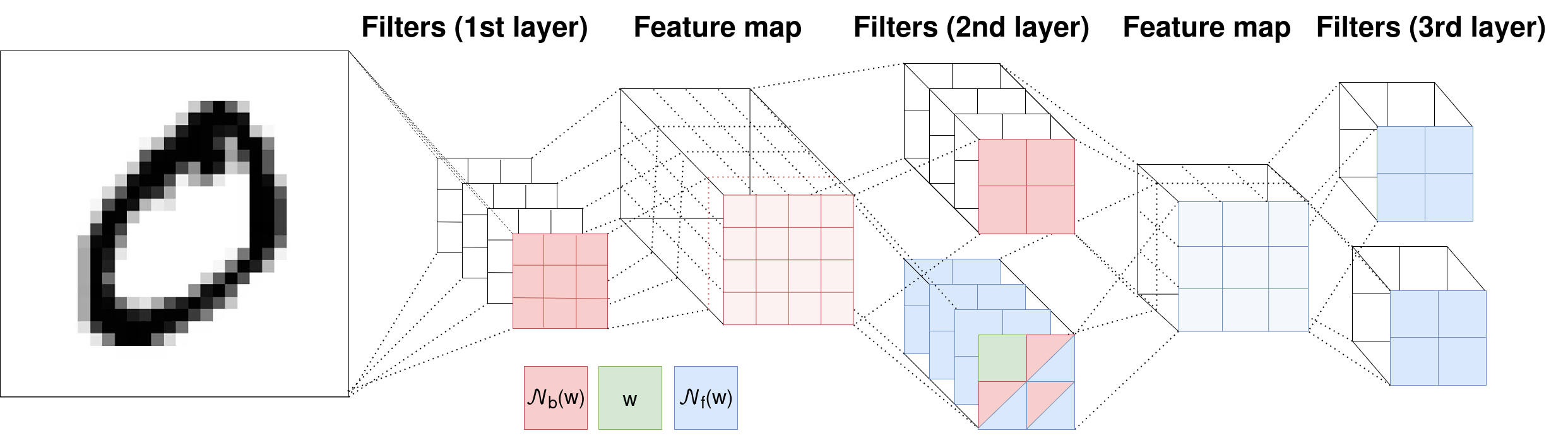}
	\caption{Diagram showing a weight $w$ and its neighborhood in an example CNN's first 3 layers. The backward neighborhood $\mathcal{N}_{b}(w)$ consists of all current filters' weights operating on the same input channel as $w$, as well as the filter from the previous layer that produces this input channel. The forward neighborhood $\mathcal{N}_{f}(w)$ consists of all weights in the current filter of $w$, as well as all filters in the next layer from the channel produced by the current layer's filter. Note that the current filters' weights operating on the same input channel as $w$ are both forward and backward neighbors, shown in both colors in the figure.}
    \label{fig:conv_neighborhood}
\end{figure*}

We construct the neighborhood of $w$ as the input to the local rule network $R$ as follows.
We start with a particular weight $w$ and its corresponding hidden state $\vec{h}$.
To give a sense of direction for the neighborhood of $w$, we define the \old{"forward"}\new{\mbox{``forward''}} neighbors and \old{"backward"}\new{\mbox{``backward''}} neighbors as those weights connected to the forward \new{(post-synaptic)} and backward \new{(pre-synaptic)} neurons of $w$; see Figure \ref{fig:pipeline} for an illustration for feedforward layers. For convolutional layers, see Figure \ref{fig:conv_neighborhood}. The forward and backward neighborhoods of $w$ are denoted as $\mathcal{N}_{f}(w)$ and $\mathcal{N}_{b}(w)$ respectively.
Weights are randomly initialized while hidden states are initialized with sinusoidal absolute positional encodings \citep{vaswani2017attention} across layers, forward neurons, backward neurons, and horizontal and vertical kernel positions (for use in convolutional networks).

\subsection{Weight Transformer}
We use a linear attention transformer to aggregate the weights and hidden states of forward and backward neighbors of an individual weight to construct a perception vector. This is passed to the MLP which updates the weight and hidden state. In the following section, we describe our implementation.

\subsubsection{Linear Attention with ELU-based Feature Map}
Consider a particular weight $w_i$ of the task network. Let $w_i$'s forward neighborhood vectors be denoted as $\mathcal{N}_{f}(w_i)$.

We apply cross-attention between $[w_i \quad h_{w_i}]$ and its forward neighborhood's weights and hidden states $\{[w_j \quad h_{w_j}] \mid j \in \mathcal{N}_{f}\}$. All operations concerning the forward neighborhood are applied to the backward neighborhood as well. Before cross-attention, we randomly drop out a percentage of whole vectors of the forward and backward neighborhood to reduce dependence on individual neighbors.
We project $[w \quad h_{w}]$ as the query vector $q$:
\begin{equation}
    q = W_{Q} \begin{bmatrix}
        w \\
        \vec{h}
    \end{bmatrix}
    \label{eq:query}
\end{equation}
\old{Applying this projection on all $w$ yields the query matrix $Q$.}\new{The full matrix of queries for all the weights in the current layer under consideration is $Q$.} Similarly for all key and value weight matrices:
\begin{equation}
\begin{array}{cc}
    K = W_{K} \begin{bmatrix}
        w_{f1} & \dots \\
        \vec{h}_{f1} & \dots
    \end{bmatrix}
    &
    V = W_{V} \begin{bmatrix}
        w_{f1} & \dots \\
        \vec{h}_{f1} & \dots
    \end{bmatrix}
\end{array}
\label{eq:keyvalue}
\end{equation}

We apply a feature map $\psi(x) = \mathrm{ELU}(x) + 1$ to the query and key matrices $Q$ and $K$ for a vector $x \in \mathbb{R}^d$ using an exponential linear unit (ELU) function \citep{clevert2015fast}.

\paragraph{Multihead Linear Attention}

We use the rotary positional encoding formulation for linear attention from \cite{roformer}, applying position-dependent rotation matrices $R$ to the feature-mapped queries and keys.
Each head of the linear attention mechanism corresponds to a particular positional dimension of a weight: layer index, forward neuron index, backward neuron index, horizontal kernel position index, and vertical kernel position index.
Kernel positions only differ for convolutional networks.
For a given attention head, let $\tilde{Q} = R\psi(Q)$ and $\tilde{K} = R\psi(K)$ denote the rotated feature-mapped queries and keys. The kernelized linear attention output is:

\begin{equation}A = D^{-1}\tilde{Q}\tilde{K}^{T}V\label{eq:attention}\end{equation}

where $D = \tilde{Q}\tilde{K}^{T}\mathbf{1} + \epsilon$ is a normalizing factor. This formulation avoids materializing the full attention matrix, scaling linearly with the number of neighbors.

The multi-head outputs are concatenated and passed through a learned output projection, followed by a residual connection, layer normalization, a two-layer feedforward network, and a second layer normalization, forming a full transformer-style block.

A forward neighborhood attention output vector $a_{f}$ constitutes our ``forward signal''. A separate attention block with its own parameters is applied to the backward neighborhood, obtaining $a_{b}$. We concatenate these with the focus weight to form the perception vector:
\begin{equation}v_{perception} = [w, \vec{h}, a_{f}, a_{b}]\label{eq:perception}\end{equation}

The updates to $w$ and $\vec{h}$ for the next timestep $t + 1$ are given by a feedforward network (the local rule head) applied to the perception vector:
\begin{equation}\Delta w^{(t+1)}, \Delta \vec{h}^{(t+1)} = R(\phi; v_{perception})\label{eq:update}\end{equation}

\begin{figure}[ht!]
  \centering
    \includegraphics[width=0.42\textwidth]{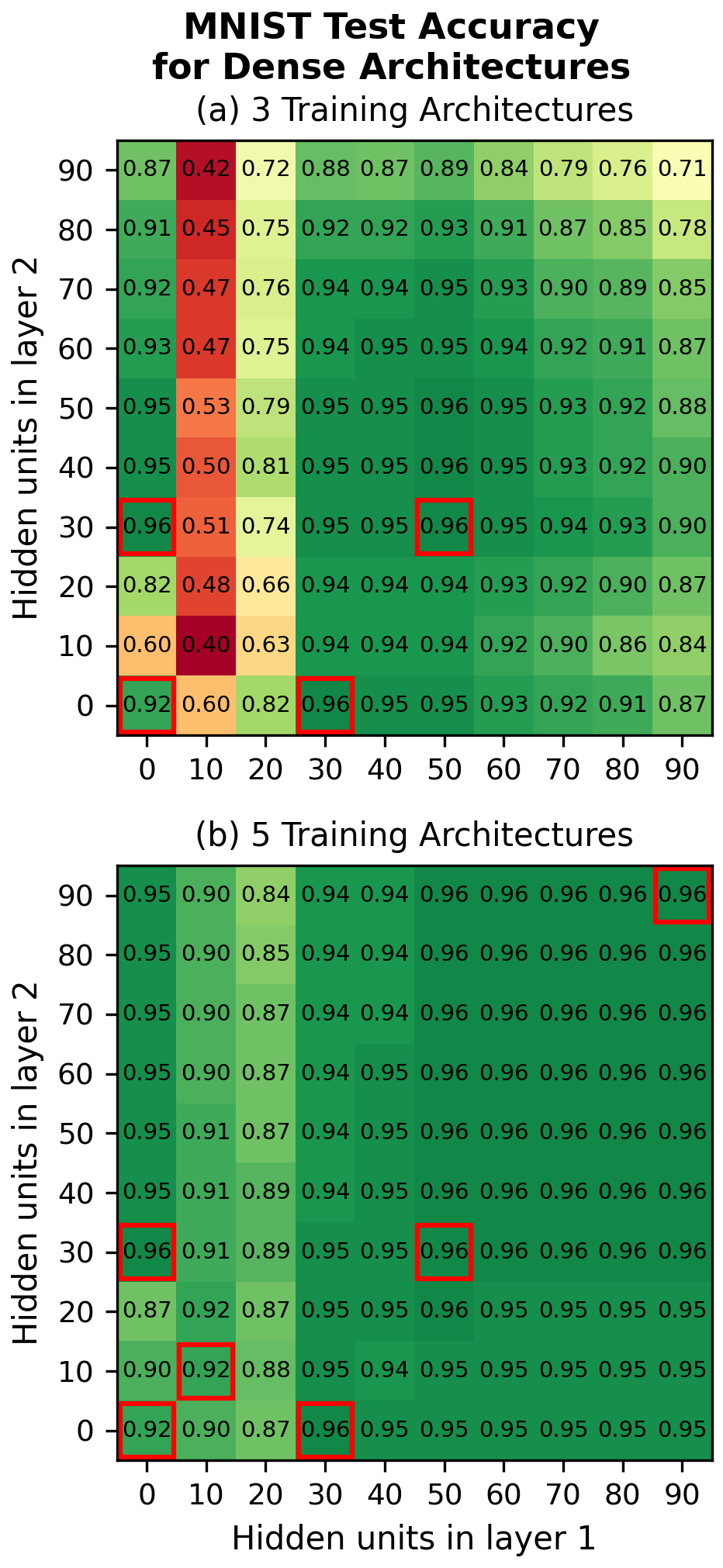}
  \caption{Heatmaps of weight transformer local rule net generating dense network architectures, with training architectures indicated with red rectangles. (a) and (b) show two settings where 3 and 5 task network architectures were used for training the local rule network. \new{The two axes are the hidden-layer widths $h_1$ and $h_2$ (width $0$ = absent layer). Configurations such as $(30,0)$ and $(0,30)$ are the same single-hidden-layer model, so one training architecture can be marked by two rectangles; this is why there appear to be more rectangles than training architectures.}}
    \label{weight-transformer-heatmap-interpolation-add-archs}
\end{figure}

\begin{figure*}[ht]
  \centering
  \includegraphics[width=\textwidth]{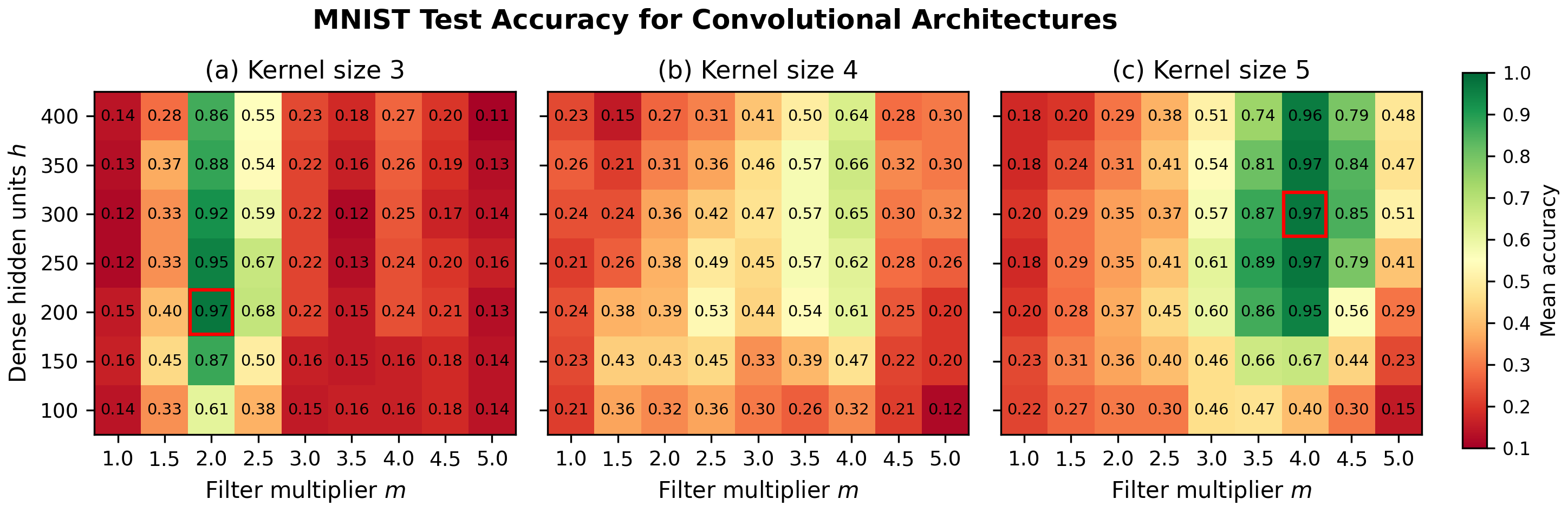}
  \caption{Heatmaps of the Weight Transformer local rule network generating convolutional task networks across three kernel sizes. Architectures use a multiplier $m$ for convolutional filters per layer (32$m$, 64$m$, 128$m$) and $h$ dense hidden units before the output layer. Training architectures are indicated with red rectangles. \new{There are two training architectures in total, one at kernel size 3 and one at kernel size 5; kernel size 4 is entirely held out.} Accuracies are on MNIST (80/20 train/test split), averaged across 5 samples per architecture.}
	\label{large-conv-grid-multiple-kernel-sizes}
\end{figure*}

\paragraph{Training the local rule neural network.}

For our experiments we use a cross-entropy classification loss for the MNIST and CIFAR100 datasets.
To train the local rule network, we first calculate the loss of the task network.
We then backpropagate through the task neural network, and the local rule neural network, over the sampled number of local rule updates of the task network, in order to calculate the gradient of the local rule network.
We use the Muon optimizer to calculate updates to the local rule network \citep{jordan2024muon}.
We only update the weights of the local rule network through backpropagation; the task network is exclusively updated by the local rule network outputs.
To simulate asynchronous updates, we apply the local rule updates stochastically to 80\% of weights of the task net per local rule update for our experiments.

\begin{algorithm}[t]
\caption{One MetaNCA update  $(t \xrightarrow{} t+1)$}
\label{alg:metanca_step}
\begin{algorithmic}[1]
\REQUIRE Task net weights $\theta$, hidden states $\mathcal{H}$, positional encodings $\mathcal{P}$, local rule $R(\phi;\cdot)$, update fraction $p$
\FOR{each layer $l = 1, \ldots, L$}
    \STATE Sample masked indices $\mathcal{M}$ with $|\mathcal{M}| = p \cdot |\theta^{(l)}|$
    \FOR{each weight index $i \in \mathcal{M}$}
        \STATE{$w \leftarrow \theta_{i}^{(l)}, \vec{h} \leftarrow \mathcal{H}_{i}^{(l)}, x_{pos} \leftarrow \mathcal{P}_{i}^{(l)}$}
        \STATE $\mathcal{N}_{f}, \mathcal{N}_{b} \leftarrow \text{GatherNeighbors}(w)$
        \STATE{\texttt{//Attention over neighbors}}
        \STATE $a_f \leftarrow \text{ForwardEncoder} ([w, \vec{h}],\; \mathcal{N}_f,\; x_{pos})$
        \STATE $a_b \leftarrow \text{BackwardEncoder}([w, \vec{h}],\; \mathcal{N}_b,\; x_{pos})$
        \STATE{\texttt{//Construct perception vector}}
        \STATE $v_{per} \leftarrow [w, \vec{h}, a_f, a_b]$
        \STATE $\Delta w, \Delta \vec{h} \leftarrow R\bigl(\phi;\; v_{per}\bigr)$
        \STATE $\theta_{i}^{(l)} \gets w + \Delta w$ \quad $\mathcal{H}_{i}^{(l)} \gets \vec{h} + \Delta \vec{h}$
    \ENDFOR
\ENDFOR
\end{algorithmic}
\end{algorithm}

\subsection{Complexity Analysis}
\label{sec:complexity}

We analyze the compute and space complexity of MetaNCA in terms of the total number of task network weights $W$, the number of local rule update steps $T$, neighborhood sizes per weight $\mathcal{N}_{f}$ and $\mathcal{N}_b$, and the local rule parameters $\phi$. We denote the compute and space complexities of a computational stage $q$ as $C_q$ and $S_q$.

Algorithm~\ref{alg:metanca_step} summarizes a single update step. For each of the $L$ layers in the task network, linear attention is computed over the local neighborhood, and the local rule MLP head produces per-weight updates $\Delta w$ and $\Delta \vec{h}$. A stochastic mask selects a fraction $p$ of weights to update each step.

Each update step has the computational cost of applying the local rule network to each weight:

\begin{align}
C_\mathrm{lr} &= \Theta\bigg (|\phi_{mlp}| \cdot W + | \phi_{attn} |\sum_{i=1}^W |\mathcal{N}^{(i)}_f| + |\mathcal{N}^{(i)}_b| \bigg)
\end{align}

The local rule updates can be parallelized across weights (and architectures), reducing wall clock time.

\paragraph{Local rule updates.} Applying $T=10$ update steps to generate a task network, the compute and space complexity is given by
\begin{align}
\label{eq:inference_time}
  C_{\mathrm{inference}} &= O(T \cdot C_\mathrm{lr}), \\
\label{eq:inference_space}
  S_{\mathrm{inference}} &= O(W \cdot h + |\phi|).
\end{align}
Memory is dominated by the task network state: each of the $W$ weights carries a hidden state and positional encoding of dimension $h$. 

\paragraph{Meta-training.} Training the local rule via backpropagation through time (BPTT) over $T$ unrolled steps introduces a constant factor for gradient computation. Gradient checkpointing\new{\mbox{ \citep{chen2016training}}} recomputes each step's forward pass during backpropagation, making memory scale with $\sqrt{T}$:
\begin{align}
\label{eq:training_time}
  C_{\mathrm{train}} &= O(T \cdot C_\mathrm{lr}) \text{ per arch.\ per batch}, \\
\label{eq:training_space}
  S_{\mathrm{train}} &= O(\sqrt{T} \cdot W \cdot h + |\phi|).
\end{align}
Training memory stores a single training task network at a time by accumulating the local rule network's gradients per task network.

\section{Experiments}

We evaluate MetaNCA on two datasets: MNIST \citep{deng2012mnist} (handwritten digit classification) and CIFAR-100 (100-class image classification). We test architecture generalization across three families of task networks:
\begin{itemize}
    \item \textbf{Dense MLPs} on MNIST: two hidden layers with 0--90 units each, forming a $[784, h_1, h_2, 10]$ architecture grid.
    \item \textbf{Convolutional networks} on MNIST: 3-layer CNNs with a filter multiplier $m$ controlling channel widths ($32m$, $64m$, $128m$), varying kernel sizes (3, 4, 5), and a dense hidden layer of $h$ units before the output. These networks range up to 2 million parameters.
    \item \textbf{ResNets} \citep{he2016deep} on CIFAR-100: 3-stage residual networks with $[1,1,1]$ blocks per stage and channel widths $[c, 2c, 4c]$, where $c$ is the base channel count.
\end{itemize}

For each experiment, we meta-train a single local rule network on a small set of task network architectures (2--5), then evaluate on a grid of architectures including both seen and unseen configurations. All reported accuracies are \old{validation}\new{test} set accuracy (80/20 train/test split), averaged over multiple samples per architecture.

\section{Results}

\subsection{Architecture Generalization: Dense MLPs}

Figure~\ref{weight-transformer-heatmap-interpolation-add-archs} shows architecture generalization for dense MLPs on MNIST. With only 3 training architectures, the local rule network achieves \old{$\geq$95\%}\new{$\geq$90\%} accuracy on most architectures with at least 30 hidden units per layer. Adding 2 more training architectures (Figure~\ref{weight-transformer-heatmap-interpolation-add-archs}b) substantially improves coverage: architectures with small layers (e.g., 10 units) that previously achieved $\sim$40--50\% now reach $\geq$90\%. \new{The two architectures added in (b) were the worst-performing configuration from the 3-architecture setting and one of the largest architectures in the grid. Because the latter sits at the boundary of the architecture space, adding it shifts this setting from primarily extrapolation toward interpolation, which partly explains the improved coverage.} \old{The largest architectures consistently match or approach training architecture performance, demonstrating strong interpolation and extrapolation across the architecture space.}\new{Performance is weakest for the smallest, lowest-capacity architectures.}

\subsection{Architecture Generalization: Convolutional Networks}

Figure~\ref{large-conv-grid-multiple-kernel-sizes} shows generalization across convolutional architectures with three kernel sizes. With only \old{two training architectures per kernel size}\new{two training architectures in total (one at kernel size 3 and one at kernel size 5)}, the local rule achieves up to 97\% accuracy on MNIST for the best configurations. Performance peaks near the training architectures and degrades gracefully for more distant configurations, confirming that the local rule learns generalizable weight-generation strategies rather than memorizing specific architectures.

\subsection{Architecture Generalization: ResNets on CIFAR-100}

\begin{figure}[ht!]
    \centering
    \includegraphics[width=0.48\textwidth]{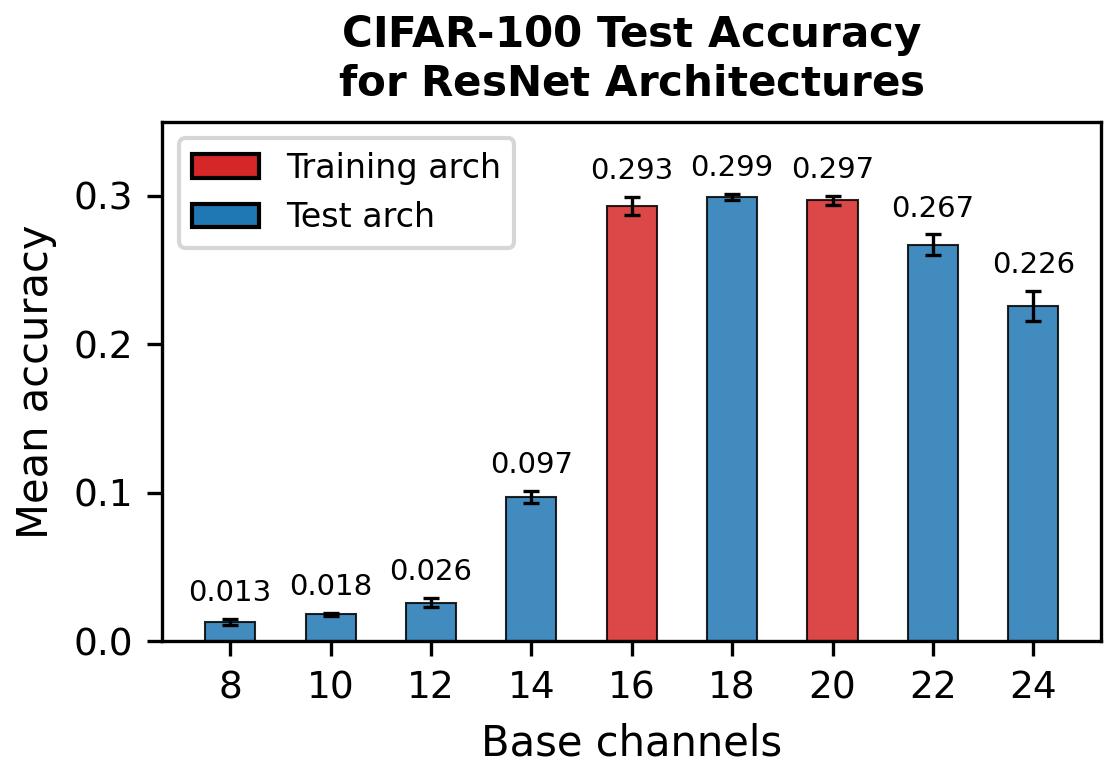}
	\caption{CIFAR-100 \old{validation}\new{test} accuracy for ResNet architectures with 3 convolutional stages (channel widths $c, 2c, 4c$) followed by global average pooling and a dense output layer, where $c$ is the base channel count. Training architectures ($c=16$, 83K params; $c=20$, 128K params) shown in red. Mean $\pm$ std over 5 samples.}
    \label{fig:cifar100resnets}
\end{figure}

We further test MetaNCA on ResNet architectures for the more challenging CIFAR-100 task (Figure~\ref{fig:cifar100resnets}). Trained on only two base channel widths ($c=16$ and $c=20$), the local rule generalizes to nearby architectures, achieving up to 29.9\% accuracy. Performance drops sharply for smaller architectures ($c \leq 14$), suggesting the local rule has learned update dynamics specific to a capacity range rather than a universal strategy.

\new{To contextualize these accuracies, we compare MetaNCA-generated networks to the same architectures trained conventionally with Adam. On MNIST the two are comparable: the $[784,50,50,10]$ network reaches 97\% with MetaNCA and 96.9\% with Adam. On CIFAR-100, MetaNCA reaches up to 29.9\%, whereas Adam-trained baselines of the same ResNets reach 41.4\% ($c{=}16$) and 42.1\% ($c{=}20$); it is possible that with more training architectures and/or hyperparameter tuning, this gap could be reduced.}

\subsection{Weight Structure}

\begin{figure*}[ht!]
    \centering
    \includegraphics[width=0.9\textwidth]{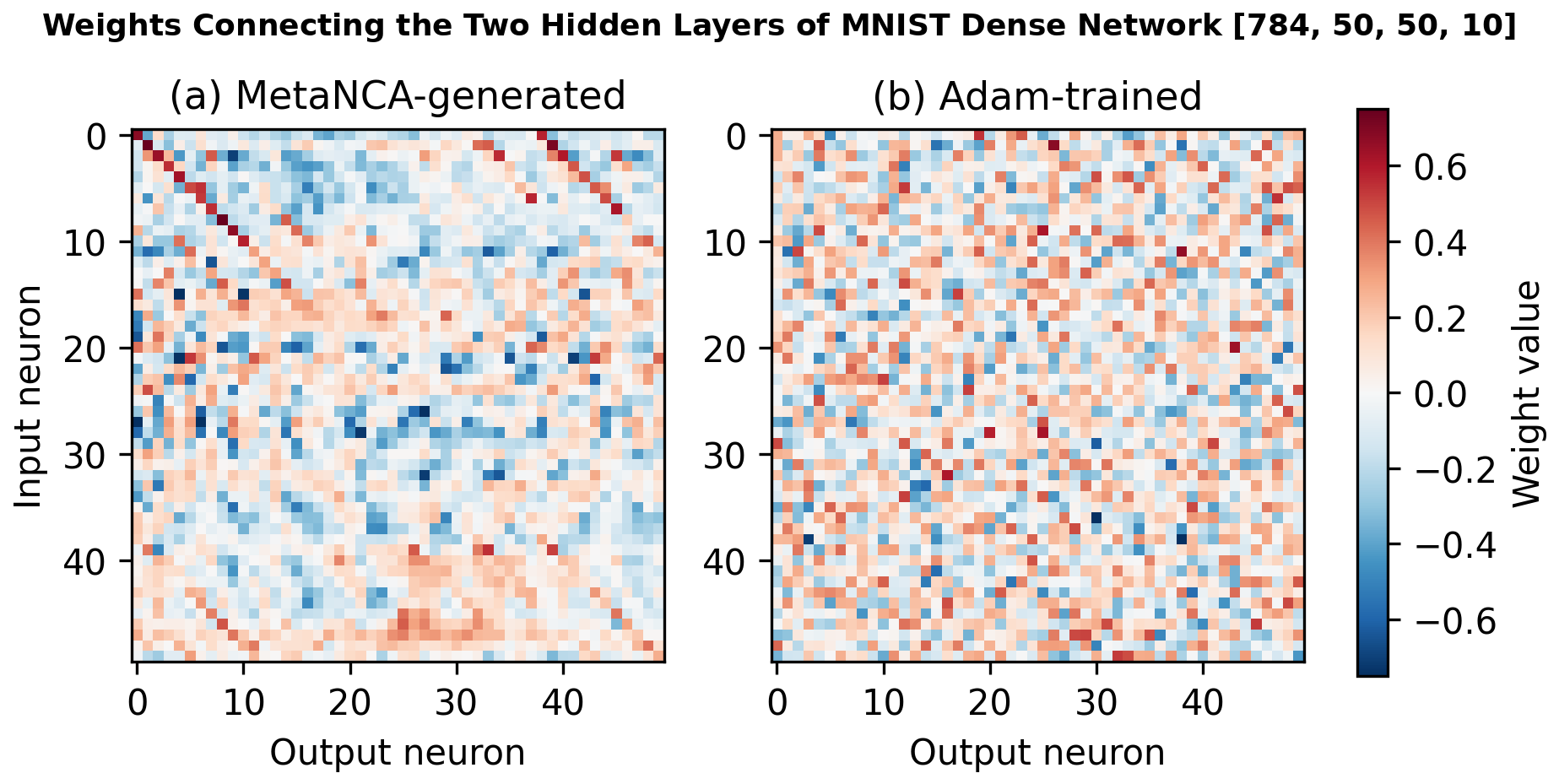}
    \caption{\old{Middle layer (50$\times$50) weight matrices}\new{Weight matrices (50$\times$50) connecting the two hidden layers} of MNIST dense networks. (a) Generated by MetaNCA after 10 local rule updates. (b) Trained with Adam for 50 epochs. Both achieve 97\% \old{validation}\new{test} accuracy.}
    \label{fig:weight-comparison}
\end{figure*}

Figure~\ref{fig:weight-comparison} compares the \old{middle layer weight matrix}\new{weight matrix connecting the two hidden layers} of a $[784, 50, 50, 10]$ MLP generated by MetaNCA (10 local rule update steps) with one trained by Adam (50 epochs). Both networks achieve similar \old{validation}\new{test} accuracy on MNIST (97\%). The MetaNCA-generated weights exhibit symmetry and more sparsity of the weights generated compared to the Adam trained weights.

\section{Discussion}
Motivated by principles of localized and distributed self-organization in biological systems, we implement a local weight-generating model to solve standard benchmarking tasks in machine learning. We show the ability of the local rules learned by MetaNCA to generalize to unseen architectures with millions of parameters. During training, the local rule learns an attractor to the high-performing task network weights from randomly initialized states.

We can view MetaNCA as compressing several training task networks into the local rule network parameters. The local rule network is a smaller description of the generation process that then decompresses into trained task networks when applied iteratively on an initialized task network. In this way, MetaNCA could be seen as an \old{architecture-agnostic}\new{architecture-flexible} model distillation method. \old{For example in the convolutional network experiment, an 80k parameter MetaNCA model generates a 2M parameter model, a compression of 4\% of the size of the task network.}\new{The rule network has a fixed size (82k parameters for MLPs and 126k for convolutional and residual networks), independent of the task network it generates. For the largest convolutional networks (2M parameters) this is approximately a $16\times$ compression; moreover, a single rule network generates an entire family of architectures rather than a single network, so its cost is amortized across the family.}

We have found that increasing the number of training architectures allows a greater degree of architecture generalization. Training on a much larger set of architectures of different types may allow for better performance on individual task networks, through transferring the capabilities of larger models to smaller ones, or different topologies allowing the learning rule to capture more useful signals for optimization.

\new{We note two limitations on the scope of these results. While the neighborhood is defined generally from a weight's forward and backward neighbors on the computation graph, instantiating it for a new architecture family (for example, the channel and kernel structure of convolutions) still requires mapping this definition onto that family's connectivity. In addition, generalization is strongest within the capacity range spanned by the training architectures and degrades for architectures far from it, as seen for the smallest ResNets. Training a single rule across multiple architecture families, rather than one rule per family as we do here, is a natural next step that the generality of the neighborhood definition should support, but which we leave to future work.}

\section{Future Work}
The method introduced here allows for fast sampling of models of architectures not seen during training.
However, the local rule updates are not conditioned directly on the data, and therefore any generalization of a local rule to new tasks would be limited at best.
In the future we wish to add a dependence of the local rule on the data, incorporating signals from activations and loss values, similar to \cite{randazzo2020MPLP} and \cite{plantec2024evolving}.

\subsection{Multitask MetaNCA}

Although it is not the focus of this current work, we have found that MetaNCA can scale to multiple concurrent tasks. By conditioning the local rule on a one-hot task vector, a single local rule network can generate task networks for MNIST and CIFAR-100 simultaneously. We found that the multitask MetaNCA produced task networks that performed comparably to those generated in the single-task setting, indicating that the local rule can learn task-specific weight-generation strategies within a shared parameter space. In future work, we aim to explore task generalization in addition to architecture generalization.


\footnotesize
\bibliographystyle{apalike}
\bibliography{bibliography}

\end{document}